# Real-time 3D Facial Tracking via Cascaded Compositional Learning

Jianwen Lou, Xiaoxu Cai, Junyu Dong, *Member IEEE*, and Hui Yu, *Senior Member, IEEE*

*Abstract*—We propose to learn a cascade of globally-optimized modular boosted ferns (GoMBF) to solve multi-modal facial motion regression for real-time 3D facial tracking from a monocular RGB camera. GoMBF is a deep composition of multiple regression models with each is a boosted ferns initially trained to predict partial motion parameters of the same modality, and then concatenated together via a global optimization step to form a singular strong boosted ferns that can effectively handle the whole regression target. It can explicitly cope with the modality variety in output variables, while manifesting increased fitting power and a faster learning speed comparing against the conventional boosted ferns. By further cascading a sequence of GoMBFs (GoMBF-Cascade) to regress facial motion parameters, we achieve competitive tracking performance on a variety of in-the-wild videos comparing to the state-of-the-art methods, which require much more training data or have higher computational complexity. It provides a robust and highly elegant solution to real-time 3D facial tracking using a small set of training data and hence makes it more practical in real-world applications.

We further deeply investigate the effect of synthesized facial images on training GoMBF-Cascade for 3D facial tracking. We apply three types synthetic images with various naturalness levels for training, and compare the performance of the tracking models trained on real data, on synthetic data and on a mixture of data. The experimental results indicate that, i) the model trained purely on synthetic facial imageries can hardly generalize well to unconstrained real-world data, ii) involving synthetic faces into training benefits tracking in some certain scenarios but degrades the tracking model's generalization ability. These two insights could benefit a range of non-deep learning facial image analysis tasks where the labelled real data is difficult to acquire.

*Index Terms*—3D facial tracking, compositional learning, boosted ferns, synthetic training imagery.

## I. Introduction

TRACKING 3D facial motion from a monocular RGB camera is a fundamental task which benefits a wide range of applications such as facial animation [1][2][3], facial reenactment [4] and emotion recognition [5]. Over the past years, a number of novel tracking algorithms [6][7][8][9][10][11][12][13] have been proposed, which led to rapid progress in this area. In particular, machine learning-based approaches that directly learn a regression function from image features to motion parameters greatly improve the tracking performance in speed, robustness and ease of use by circumventing the compute-intensive online optimization steps and leveraging a high-quality training corpus. Whereas the current state-of-the-art can deliver impressive tracking results even for very challenging cases such as large facial pose [2][11] and severe occlusion [8], the regression algorithms they applied are still not effective in dealing with multi-modal motion parameters.

Facial motion parameters such as those of head pose and expression vary significantly in scale and have different influences on facial geometry. Accurately regressing to multi-modal motion parameters from image features is a challenging task. Its learning process is prone to focusing more on parameters (e.g. 2D landmark displacements) with higher dimensionality and larger magnitude, while neglecting those (e.g. rotation angles) that impact heavily on facial geometry but with smaller magnitude. To solve this problem, previous methods either carefully chose weights to balance the parameter effects on feature selection when training a boosted ferns [6][14] or minimized a more complicated photo-geometric difference loss instead of the parameter difference loss when training a convolutional neural network [11][13]. For the first method, the process of finding appropriate weights is somewhat clumsy and it's arguable if those empirical weights can correctly reflect the parameter's significance. The latter method embeds parameter effects into the gradient of the loss function but at high computational cost.

To tackle the aforementioned problems, we adopt the compositional learning framework and propose a novel boosting method that can efficiently cope with the modality variety in output variables in this paper. The proposed method first learns a modular boosted ferns [15] which is a shallow composition of several independent regression models with each is a boosted ferns trained targeting only partial output variables of the same modality. All fern leaves are then simultaneously optimized by minimizing a global loss function defined on all output variables, which can be solved efficiently with Ridge Regression [16]. The complementary information between the old biased ferns is thus injected into the refined fern leaves, producing a new boosted ferns which is a deep composition of the pre-learnt modality-specific regression models and has much stronger predictive power. We call this method Globally-optimized Modular Boosted Ferns - GoMBF.

This work was supported in part by the ************.

Jianwen Lou, Xiaoxu Cai and Hui Yu are with the School of Creative Technologies, University of Portsmouth, Portsmouth, PO1 2DJ, UK.

Junyu Dong is with the School of Information Science and Engineering, Ocean University of China, Qingdao, 266100, China.

Corresponding Author: Hui Yu, hui.yu@port.ac.uk.

As in [15][17], we then build facial motion regression with a cascade of GoMBFs (GoMBF-Cascade) which progressively update motion parameters from an initial state by calling GoMBF to estimate an increment stage-by-stage. Extensive experiments on in-the-wild videos demonstrate that GoMBF is superior in both the fitting power and the learning speed comparing against the traditional boosted ferns that has been widely applied in 2D/3D facial shape regression [1][2][15]. The resulting GoMBF-Cascade regression delivers competitive 3D facial tracking performance comparing to the state-of-the-art methods [10][11] which require much more training data or have a much higher computational complexity.

Along with a reliable regression algorithm, quality training data is another key factor to the tracking model's robustness. For 3D facial tracking, the training data typically means facial images paired with the ground truth 3D geometry. Such data is normally acquired by multi-view stereo [9][18], photometric [11] or 2D landmark-based [2] reconstruction which requires either complicated and expensive multi-camera setups or laborious manual annotations. Alternatively, synthetic generation of training imagery provides a more economic and efficient data collection way. This approach has shown effectiveness on training deep convolutional neural networks for accurate 3D facial tracking and reconstruction in recent studies [11][19][20]. However, it remains unclear whether the synthetic data also works on training non-deep learning methods such as GoMBF-Cascade. In this study, we explore this question via progressively adjusting the naturalness of synthetic images for training GoMBF-Cascade and comparing between tracking models that are trained on real data, on synthetic data and on a mixture of data. In our experiments, the GoMBF-Cascade models trained purely on synthesized images have shown poor tracking performance on real videos and become more biased after incorporating the synthetic data into training.

In summary, our main contributions are as follows:

1) Based on compositional learning, we develop a novel boosting algorithm – GoMBF which deals effectively with the modality variety in output variables. GoMBF shows stronger fitting power and a faster learning speed when comparing with the conventional boosted ferns [1][2][15]. It can be seamlessly adapted to any other multi-output regression tasks in theory.

2) By cascading GoMBFs for facial motion regression, we obtain a competitive 3D facial tracking performance compared with the state-of-the-art methods [10][11], which rely on large-scale training data or bear much higher computational complexity. It thus offers a robust and very practical solution to real-time 3D facial tracking.

3) We carry out an in-depth investigation into the effect of synthetic data on training GoMBF-Cascade for 3D facial tracking, which provides a novel view of the synthetic data's role in training non-deep learning method for facial image analysis where the real labelled data is difficult to obtain.

## II. RELATED WORK

Real-time 3D facial motion capture from a monocular RGB video has been extensively studied in computer graphics and vision communities. It is normally achieved by estimating a group of parameters which encode facial expression and head pose within a low-dimensional space from video frames. Generally, there are two types of approaches to estimate those parameters - optimization-based approach and learning-based approach, which divides the existing studies into two main streams. In this section, we will review the most relevant works from the two categories and also discuss how the synthetic data has been used in learning-based approaches. For a more comprehensive review on related topics, interested readers are directed to [21].

*A. Optimization-based Approaches*

Optimization-based approach is built upon the idea of analysis-by-synthesis where a parametric face model is iteratively adapted until the synthesized face matches the target image. It is formulated as minimizing a highly non-linear objective function which typically enforces alignment on sparse/dense feature points [1][7][22] and pixel intensities [4][10] between the synthesized result and the input data, while regularizing the estimated shape parameters to lie within a valid range for generating a plausible face. Solving this optimization problem usually requires massive computing power such as GPU acceleration to achieve real-time performance [4][10]. This hinders the approach's deployment to platforms with limited computing resources.

*B. Learning-based Approaches*

Learning-based approach bypasses the costly optimization step by estimating facial motion parameters from image features through a regression learned from a hand-picked training corpus. Cao et al. [2][14] pioneered this area by employing a two-level boosted regression – Explicit Shape Regression (ESR) [15] to map facial appearance features to motion parameters. Their method was trained on public image datasets with estimated 3D facial data and achieved impressive tracking performance on in-the-wild videos. The work opened up a new era of learning-based 3D facial tracking and motivated a bunch of follow-ups [3][6][8] which extended the tracking to more challenging cases such as capturing facial geometry details (e.g. wrinkles and dimples) [3] and tracking under severe occlusions [8]. Despite the great success achieved by these works, the boosted ferns employed in ESR is deficient in handling the modality variety of motion parameters whose scale and influence on facial geometry differ a lot from each other. To mitigate this problem, a few studies [6][14] applied a weighting-vector to balance the parameter effects on feature selection in fern learning. This intuitive strategy is moderately inefficient and it's doubtful if those empirical weights can fully reflect the parameter's significance. More recent studies [11][13] instead employed a deep convolutional neural network coupled with a photo-geometric difference loss to learn the facial motion regression. This method inherently incorporates the motion parameter's influence on facial geometry into the gradient of the loss function, which however bears a high computational complexity. Alternatively, the proposed GoMBF first learns an exclusive boosted ferns for each kind of motion parameters and then optimizes all fern leaves towards the whole regression target with linear regression, which explicitly handles the output variable's modality variety in a fairly efficient manner.

*C. Learning from Synthetic Data*

In contrast to traditional 3D facial data harvesting methods which need multi-camera setups or manual annotations, synthetic generation of training imagery offers a highly efficient and economic data collection way. Learning from synthetic data is attracting more and more attention in 3D facial tracking and reconstruction [11][19][20]. Richardson et al. [19] proposed to render photo-realistic 3D facial meshes and images using 3D Morphable Model (3DMM) [23] and Phong illumination [24] for training a convolutional neural network (CNN) for 3D face reconstruction. Though the network was trained purely on synthetic data, it generalized well to real-world face images. Guo et al. [11] later used albedo and lighting coefficients inferred from real face images to render more natural-look faces for training the CNN. Their model achieved high-quality tracking results on in-the-wild videos. A more recent study [25] shows that priming deep networks by pre-training them with synthetic faces is helpful, e.g. it can reduce the negative effects of the training data bias. Whereas there is continuous evidence manifesting that the synthesized faces favour deep learning methods, it remains unclear if such data also benefits non-deep learning methods. To our knowledge, only McDonagh and his colleagues [6] have succeeded in learning a boosted ferns from the synthesized faces for personalized 3D facial tracking. However, their synthetic generation of training imagery was based on a high-quality facial rig of the user's face obtained from an offline capture system and a simulated illumination driven by light probe data acquired at the target environment. This process can hardly be adapted to unconstrained facial tracking where the target environment is unknown in the training phase. In this paper, we provide a novel view of the synthetic data's role in training non-deep learning methods by incorporating three kinds of synthetic data for training our GoMBF-Cascade and comparing tracking models trained on real data, on synthetic data and on a mixture of data.

## III. METHOD OVERVIEW

This section overviews our 3D facial tracking framework. We first introduce the parametric face model for representing the facial shape, then formulate the tracking workflow which is driven by the proposed GoMBF-Cascade motion regression.

*A. Parametric Face Model*

A 3D facial mesh is typically formed with a vector of stacked vertex coordinates $S = [x_1, y_1, z_1, ..., x_n, y_n, z_n]^T$ ($n = 53,215$ in this paper) and a predefined connectivity. The lengthy coordinate vector can be calculated as a weighted sum of a few basis vectors, which leaves weights the only control parameters

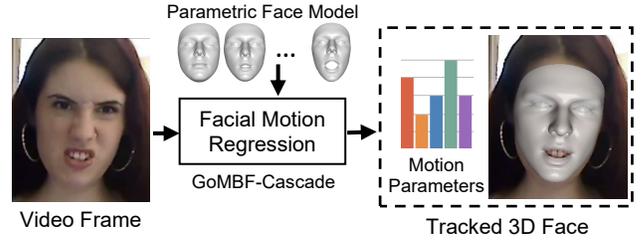

Fig. 1. 3D facial tracking workflow.

and generates a low-rank representation of the facial mesh:

$$S = B_{id}\boldsymbol{\alpha} + B_{exp}\boldsymbol{\delta} \tag{1}$$

As shown in Eq. (1), $B_{id} = \left[\mathbf{b}_0^{id}, \mathbf{b}_1^{id} ..., \mathbf{b}_{m_{id}}^{id}\right]$ is the linear basis for representing facial identity, in which $\mathbf{b}_0^{id}$ is the mean face in neutral expression. $\boldsymbol{\alpha} = \left[1, \alpha_1, ..., \alpha_{m_{id}}\right]^T$ denotes the relevant identity coefficients. $B_{exp} = \left[\mathbf{b}_1^{exp}, ..., \mathbf{b}_{m_{exp}}^{exp}\right]$ is composed of delta blendshapes of the mean face $\mathbf{b}_0^{id}$ for representing facial expression, whose coefficients - $\boldsymbol{\delta} = \left[\delta_1, ..., \delta_{m_{exp}}\right]^T$ are bounded between 0 and 1. We get $B_{id}$ ($m_{id} = 80$ as only the first 80 principal components are used in this paper) from the Basel Face Model (BFM) [23] and generate $B_{exp}$ ($m_{exp} = 46$) from FaceWarehouse [26] using deformation transfer [27].

To map $S$ which is measured in world space to image space, we apply an ideal pinhole camera model. Given a 3D point $\mathbf{v}$ in $S$, the corresponding 2D image point $\mathbf{p} = [p_x, p_y]^T$ can be obtained as:

$$\mathbf{p} = \Pi_{\mathbf{Q}}(\mathbf{R}\mathbf{v} + \mathbf{t}) \tag{2}$$

where $\mathbf{R}$ is the rotation matrix parameterized by Euler angles (*yaw*, *pitch* and *roll*) $\boldsymbol{\theta} \in \mathbb{R}^3$, and $\mathbf{t} \in \mathbb{R}^3$ is the translation vector. $\Pi_{\mathbf{Q}}$ denotes a perspective projection operator parameterized by $\mathbf{Q} = [f, 0, u_0; 0, f, v_0; 0,0,1]$ in which $f$ is the focal length and $(u_0, v_0)$ is the image centre. In practice, the estimated 3D face - $S$ and the camera model - $\{\mathbf{Q}, \mathbf{R}, \mathbf{t}\}$ may not fully match the face image. To compensate for this discrepancy, we follow [2] by using a 2D landmark displacement vector $\mathbf{D} \in \mathbb{R}^{132}$ to add onto the projected landmark coordinates to acquire 66 more accurate landmarks on the image.

The combination of parameters - $\{\boldsymbol{\alpha}, \boldsymbol{\delta}, \mathbf{Q}, \boldsymbol{\theta}, \mathbf{t}, \mathbf{D}\}$ provides a compact representation of both the 3D and 2D facial shapes. $\boldsymbol{\alpha}$ and $\mathbf{Q}$ are invariant across the whole video sequence for the same human subject. $P = [\boldsymbol{\delta}; \boldsymbol{\theta}; \mathbf{t}; \mathbf{D}] \in \mathbb{R}^{184}$ controls facial motion and changes frame by frame.

*B. Tracking Workflow*

Based on the parametric face model, 3D facial tracking from a monocular RGB video can be casted into regressing motion parameters $P$ from a video frame $I$ (see Fig. 1):

$$P = \mathcal{R}(I, \boldsymbol{\alpha}, \mathbf{Q}, P^0) \tag{3}$$

where $\mathcal{R}(\cdot)$ is the regression function, $P^0$ denotes the initial motion parameters generated from the previous frame's estimation for enforcing temporal coherence. We build $\mathcal{R}(\cdot)$ by learning a linear sequence of GoMBFs (GoMBF-Cascade) which gradually refines $P$ from $P^0$ to fit with the current frame. $\boldsymbol{\alpha}$ and $\mathbf{Q}$ are estimated from the first frame and keep fixed for the remaining frames.

## IV. Facial Motion Regression with GoMBF-Cascade

This section starts with introducing boosted ferns [15] which is the building block of our facial motion regression method. Then we elaborate the proposed globally-optimized modular boosted ferns - GoMBF and GoMBF-Cascade regression.

### A. Boosted Ferns

**Prediction.** Boosted ferns [15] is an ensemble of ferns, each fern addresses the residual of the regression target left by the preceding ferns. Its prediction is therefore the sum of all ferns' outputs. Fern is a particular instance of decision tree, which applies an identical node-splitting test for all nodes at the same tree level. The prediction of a fern with $F + 1$ levels can be formulated in a compact form:

$$\mathbf{y} = \mathbf{w}\phi(\mathbf{x}) \tag{4}$$

where $\mathbf{w}$ is a matrix of $2^F$ columns with each column stores a leaf node's prediction of output variables, $\phi(\cdot)$ represents the fern's structure (the learned node-splitting tests) which maps the data sample $\mathbf{x}$ to a one-hot vector of $2^F$ rows with each row indicating if $\mathbf{x}$ falls inside a leaf node or not (1 for yes, 0 for no), and $\mathbf{y}$ is the fern's prediction of $\mathbf{x}$. The prediction of a boosted ferns (see Fig. 2a) with $K$ ferns is thus:

$$\mathbf{y} = \sum_{i=1}^{K} \mathbf{w}_i \phi_i(\mathbf{x}) = W\Phi(\mathbf{x}) \tag{5}$$

where $W = [\mathbf{w}_1, ..., \mathbf{w}_K]$ and $\Phi(\mathbf{x}) = [\phi_1(\mathbf{x}); ...; \phi_K(\mathbf{x})]$ which is highly sparse.

**Training.** Training a boosted ferns equals to progressively training a sequence of ferns, where each fern's training loss is defined on the residual of the regression target. Specifically, a fern with $F + 1$ levels is built in two consecutive steps:

1) Learn the mapping function - $\phi(\cdot)$. It is to learn a series of node-splitting tests, each for sending a data sample $\mathbf{x}$ to the right child node if the test is satisfied or to the left child node if not. Typically, a node-splitting test is about selecting a feature from $\mathbf{x}$ and comparing it to a threshold. As in [15], we calculate the differences (referring to image pixel differences in our case) between $\mathbf{x}$'s elements and select the one that has the highest Pearson Correlation with a random projection (generated from a Gaussian distribution) of the regression target as the feature for splitting the node. A threshold is then randomly sampled from a uniform distribution which is scaled by the selected feature's maximum absolute value in the training set [15]. After repeating the process of feature selection and threshold sampling $F$ times, we can obtain the fern's $\phi(\cdot)$.

2) Learn the leaf matrix - $\mathbf{w}$. With the learned $\phi(\cdot)$, all training samples can be sent level by level from the fern root all the way down to one of the $2^F$ leaf nodes. For each leaf node, we acquire its prediction of output variables by averaging the regression targets of all training samples falling inside this node with a shrinkage to overcome overfitting [15] and save it into the corresponding column of $\mathbf{w}$.

### B. Globally-optimized Modular Boosted Ferns

Whereas boosted ferns has been successfully applied in 2D/3D shape regression [1][2][6][14][15], we found it has limitations when regressing to multi-modal output variables such as facial motion parameters $P = [\boldsymbol{\delta}; \boldsymbol{\theta}; \mathbf{t}; \mathbf{D}]$. As shown above, the prediction of a boosted ferns relies heavily on the node-splitting features. The aforementioned correlation-based feature selection method can efficiently learn good features when output variables are of a single modality. However, if output variables such as motion parameters contain multiple modalities, it is prone to selecting features that are more discriminative to output variables (e.g. 2D landmark displacements $\mathbf{D}$) with higher dimensionality and larger magnitude, while less informative to those (e.g. rotation angles $\boldsymbol{\theta}$) which are relatively negligible in numerical scale but significant in semantics. This severely degrades boosted fern's fitting power. To solve the problem, we follow the compositional learning framework and propose a globally-optimized modular boosted ferns – GoMBF, which is built in two consecutive phases (see Fig. 2b): 1) learn a modular boosted ferns in which each module regresses partial output variables of the same modality; 2) optimize all fern leaves towards the whole regression target by solving a linear regression.

*1) Learning a Modular Boosted Ferns*

A modular boosted ferns is a shallow composition of multiple regression models with each is also a boosted ferns trained independently for regressing partial output variables of the same modality, which refers to the increment of a kind of motion parameters in our case (the number of ferns is $K_{\boldsymbol{\delta}}$, $K_{\boldsymbol{\theta}}$, $K_{\mathbf{t}}$, $K_{\mathbf{D}}$ respectively):

$$\begin{cases} [\Delta\boldsymbol{\delta}; \mathbf{0}; \mathbf{0}; \mathbf{0}] = W_{\boldsymbol{\delta}}\Phi_{\boldsymbol{\delta}}(\mathbf{x}) \\ [\mathbf{0}; \Delta\boldsymbol{\theta}; \mathbf{0}; \mathbf{0}] = W_{\boldsymbol{\theta}}\Phi_{\boldsymbol{\theta}}(\mathbf{x}) \\ [\mathbf{0}; \mathbf{0}; \Delta\mathbf{t}; \mathbf{0}] = W_{\mathbf{t}}\Phi_{\mathbf{t}}(\mathbf{x}) \\ [\mathbf{0}; \mathbf{0}; \mathbf{0}; \Delta\mathbf{D}] = W_{\mathbf{D}}\Phi_{\mathbf{D}}(\mathbf{x}) \end{cases} \tag{6}$$

where $\Delta\boldsymbol{\delta}$, $\Delta\boldsymbol{\theta}$, $\Delta\mathbf{t}$ and $\Delta\mathbf{D}$ denote the predictions of motion parameter increments, $\mathbf{0}$ represents the zero vector with variant rows for extending the left output vector to match $P$'s size. Equation (6) can be written in a more compact form:

$$\Delta P = W_P \Phi_P(\mathbf{x}) \tag{7}$$

where $\Delta P = [\Delta\boldsymbol{\delta}; \Delta\boldsymbol{\theta}; \Delta\mathbf{t}; \Delta\mathbf{D}]$, $W_P = [W_{\boldsymbol{\delta}}, W_{\boldsymbol{\theta}}, W_{\mathbf{t}}, W_{\mathbf{D}}]$ and $\Phi_P(\mathbf{x}) = [\Phi_{\boldsymbol{\delta}}(\mathbf{x}); \Phi_{\boldsymbol{\theta}}(\mathbf{x}); \Phi_{\mathbf{t}}(\mathbf{x}); \Phi_{\mathbf{D}}(\mathbf{x})]$. The method reduces the original difficult regression task to four simpler sub-tasks which require a very small number of ferns for each sub-task and can be solved efficiently using parallel programming. This in turn avoids the interference from the output variable's modality variety on feature selection during fern learning.

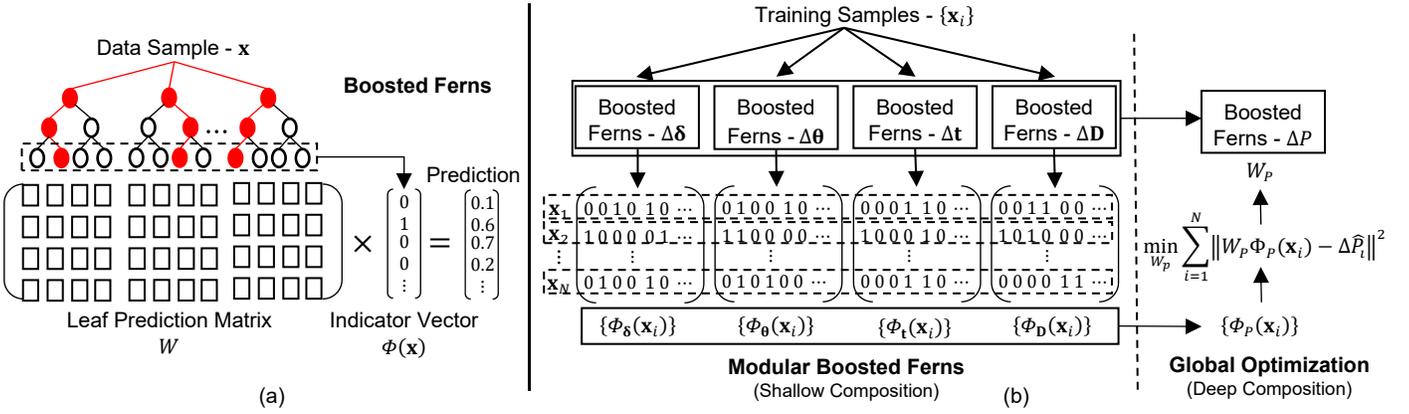

Fig. 2. Illustration of the boosted ferns and the GoMBF built with compositional learning: (a) boosted ferns, (b) GoMBF.

*2) Global Optimization*

Due to the nature of modular boosted ferns, each module learns features biased towards partial output variables of a specific modality. Those features are complementary to each other, e.g. the features that are discriminative in estimating facial expression $\boldsymbol{\delta}$ could also benefit the estimation of 2D landmark displacements $\mathbf{D}$ as both parameters encode non-rigid facial motion. However, such complementary information between inter-modular ferns has not been exploited when making prediction in Eq. (7). For example, $W_P$'s $W_\delta$ only contributes to predicting expression coefficients $\boldsymbol{\delta}$, while remaining idle when predicting the other three motion parameters. It makes the compositional regression model loosely articulated and less optimal. To this end, we propose to optimize all the pre-learnt fern leaves - $W_P$ by minimizing a common objective function defined on the whole regression target:

$$\min_{W_P} \sum_{i=1}^{N} \left\| W_P \Phi_P(\mathbf{x}_i) - \Delta \widehat{P}_i \right\|^2 \qquad (8)$$

where $N$ is the number of training samples, $\Delta \widehat{P}_i$ is the regression target of sample $i$. Equation (8) is the well-known linear least squares problem which can be solved efficiently with Ridge Regression [16]. After updating Eq. (7) with the new $W_p$, we obtain a globally-optimized modular boosted ferns – GoMBF, which is a deep composition of the pre-learnt modality-specific regression models. In our 3D facial tracking experiments, GoMBF has shown stronger fitting power and a faster learning speed than the conventional boosted ferns [15]. Moreover, it can be seamlessly applied to any other multi-output regression tasks in theory.

It is worth pointing out that GoMBF has conceptual links with two existing methods to some degree, which were developed for face alignment [28] and Random Forest refinement [29]. However, GoMBF is fundamentally different from those two methods in the following two main aspects: 1) GoMBF is designed to deal with the modality variety in regression output variables, while [28] is to learn discriminative local texture features for robust 2D landmark detection and [29] is to fill the gap between the training and the testing of Random Forest. 2) GoMBF is based on boosted ferns (boosting), while both [28] and [29] were based on Random Forest (bagging).

*C. GoMBF-Cascade Regression*

Following the basic idea of cascaded regression which has shown robustness in various shape regression tasks [15][17][28], we frame our facial motion regression with a cascade of GoMBFs, which we name as GoMBF-Cascade (see Fig. 3). For a video frame $I$, beginning with an initial motion vector $P^0$, GoMBF-Cascade gradually refines $P$ by calling GoMBF to estimate a motion increment $\Delta P^t$ stage-by-stage:

$$\begin{aligned} P &= P^0 + \sum_{t=1}^{T} \Delta P^t \\ \Delta P^t &= W_P^t \Phi_P^t(\mathbf{x}^t) \\ \mathbf{x}^t &= \mathcal{F}^t(I, \boldsymbol{\alpha}, \mathbf{Q}, P^{t-1}) \end{aligned} \qquad (9)$$

where $T$ is the number of stages, $W_P^t$ and $\Phi_P^t(\cdot)$ represent the GoMBF learned at stage $t$. $\mathbf{x}^t$ is a vector of pixel intensities extracted from image $I$ by $\mathcal{F}^t(\cdot)$ for representing the appearance of the facial shape - $\{\boldsymbol{\alpha}, \mathbf{Q}, P^{t-1}\}$ output from stage $t-1$. It can be found that Eq. (9) is a detailed expansion of Eq. (3). In the following, we will explain in detail the training, runtime prediction and appearance vector extraction of the GoMBF-cascade regression.

*1) Training*

To train the regression, we first create guess-truth motion parameter pairs $\{\widehat{P}_i, P_{ij}^0\}$ for each training image $I_i$. The guess-truth pairs simulate the runtime situation where facial motion between two adjacent video frames is assumed to be small. Specifically, given a facial image $I_i$ and its ground-truth facial motion parameters $\widehat{P}_i$, we set the initial 2D landmark displacements as zeros and perturb along $\widehat{P}_i$'s three other dimensions - $\widehat{\boldsymbol{\delta}}_i, \widehat{\boldsymbol{\theta}}_i, \widehat{\mathbf{t}}_i$ with random noise to get several guesses $\{P_{ij}^0\}$ of the initial motion parameters $P_i^0$:

- *Random Expression.* $P_{ij}^0 = [\boldsymbol{\delta}_{ij}^0; \widehat{\boldsymbol{\theta}}_i; \widehat{\mathbf{t}}_i; \mathbf{0}]$, where $\boldsymbol{\delta}_{ij}^0 = \widehat{\boldsymbol{\delta}}_{i'}$ ($i' \neq i$) is the ground-truth expression coefficients of image $I_{i'}$ which is randomly chosen from the training set.
- *Random Rotation.* $P_{ij}^0 = [\hat{\boldsymbol{\delta}}_i; \boldsymbol{\theta}_{ij}^0; \hat{\mathbf{t}}_i; \mathbf{0}]$, where $\boldsymbol{\theta}_{ij}^0 = \hat{\boldsymbol{\theta}}_i + \Delta\boldsymbol{\theta}_{ij}$. $\Delta\boldsymbol{\theta}_{ij}$ is composed of random Euler angles sampled from three independent normal distributions.
- *Random Translation.* $P_{ij}^0 = [\hat{\boldsymbol{\delta}}_i; \hat{\boldsymbol{\theta}}_i; \mathbf{t}_{ij}^0; \mathbf{0}]$, where $\mathbf{t}_{ij}^0 = \hat{\mathbf{t}}_i + \Delta\mathbf{t}_{ij}$. $\Delta\mathbf{t}_{ij}$ is a random translation vector whose elements are sampled from three independent normal distributions.

For each training image, we generate 30 guess-truth pairs for the random expression category and 8 pairs for each of the other two categories.

After constructing the set of $\{I_i, \boldsymbol{\alpha}_i, \mathbf{Q}_i, \hat{P}_i, P_{ij}^0\}$, the GoMBF-cascade regression is trained in $T$ stages. In each stage, we extract facial shape appearance vectors from all training images $\{I_i\}$ with a pre-built $\mathcal{F}^t(\cdot)$ and learn a GoMBF - $\{W_P^t, \Phi_P^t(\cdot)\}$ following the procedure explained in *Part A* and *Part B*.

*2) Runtime Prediction*

For the first video frame, we locate the face using the Viola-Jones detector [30] and detect 66 landmarks with a pre-trained SDM [31] model. Then, we predict its camera and facial shape parameters - $\{\boldsymbol{\alpha}, \mathbf{Q}, P\}$ by fitting the aforementioned parametric face model to the detected 2D landmarks, which is achieved by minimizing the following energy with the coordinate-descent method:

$$E = E_{lan} + E_{reg} \quad (10)$$
$$E_{lan} = \sum_{k=1}^{66} \left\| \mathbf{\Pi}_\mathbf{Q}\left(\mathbf{R}(B_{id}\boldsymbol{\alpha} + B_{exp}\boldsymbol{\delta})^{(l_k)} + \mathbf{t}\right) - \mathbf{p}_d^{(k)} \right\|^2$$
$$E_{reg} = w_1 \sum_{i=1}^{80}\left(\frac{\alpha_i}{\sigma_i}\right)^2 + w_2 \sum_{i=1}^{46}|\delta_i|$$

where $E_{lan}$ represent the landmark fitting error and $E_{reg}$ is the regularization term to enforce $\boldsymbol{\alpha}$ to stay statistically close to the mean and $\boldsymbol{\delta}$ to be sparse. In $E_{lan}$, $\mathbf{p}_d^{(k)}$ is the position of the $k$th detected 2D landmark and $(B_{id}\boldsymbol{\alpha} + B_{exp}\boldsymbol{\delta})^{(l_k)}$ extracts the corresponding $l_k$th vertex on the 3D facial mesh. In $E_{reg}$, $\sigma_i$ is the standard deviation of $\alpha_i$, $w_1$ and $w_2$ balance the two sub-objectives. We set $w_1$ and $w_2$ as 10 and 1 respectively. For $\mathbf{Q}$, we set the focal length $f$ as 1,000 and the principal point as the image center. This simple strategy is proven to be effective in our experiment. We then solve for $\boldsymbol{\alpha}$ and $\boldsymbol{\delta}$ by applying the L-BFGS-B solver [32] to constrain $\boldsymbol{\delta}$'s elements to lie within [0,1], and find the rigid facial motion $\{\mathbf{R}, \mathbf{t}\}$ using the POSIT algorithm [33]. The energy converges in three iterations. After each iteration, we update the indices $\{l_k\}$ of contour landmarks on the facial mesh as in [34]. Once we had the estimations of $\{\boldsymbol{\alpha}, \boldsymbol{\delta}, \mathbf{Q}, \boldsymbol{\theta}, \mathbf{t}\}$, $\mathbf{D}$ can be obtained by subtracting the projected 2D landmark positions as computed in Eq. (2) from the detected 2D landmark positions.

For each subsequent frame, we initialize its motion parameter $P$ based on the estimation $P_{prev} = [\boldsymbol{\delta}_{prev}; \boldsymbol{\theta}_{prev}; \mathbf{t}_{prev}; \mathbf{D}_{prev}]$ of the previous frame and call the learned GoMBF-Cascade regression to update $P$ to align with the current facial shape. Specifically, we initialize $P$ with $\boldsymbol{\theta}_{prev}$ and $\mathbf{t}_{prev}$, and set it's $\mathbf{D}$ as zeros. For facial expression, we found that directly inheriting

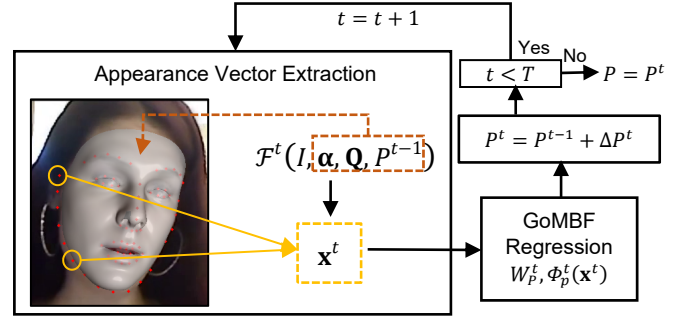

Fig. 3. The pipeline of GoMBF-Cascade facial motion regression.

$\boldsymbol{\delta}_{prev}$ for initialization will lead to implausible expression estimation and the error will accumulate across frames. This is probably due to the non-rigid nature of facial expression which makes the distribution of expression coefficients complex and difficult to be covered by the training set with limited samples. To solve the problem, we select from the training set the expression coefficients that are closest to $\boldsymbol{\delta}_{prev}$ to initialize $P$. The distance between two expression coefficient vectors is measured as the mean average distance of landmarks extracted from the corresponding 3D facial meshes. In practice, we apply multiple initial $P$s for regression and take the mean of all the outputs as the final prediction. Those initial $P$s are generated with $\boldsymbol{\delta}_{prev}$'s $L$ closest expression vectors in the training set. With the newly predicted facial motion parameters, we update the indices of contour landmarks on the facial mesh as in [34].

*3) Appearance Vector Extraction*

As shown in Eq. (9), instead of directly sending the image into the regressor, a pixel intensity vector for representing facial shape appearance is extracted from the image by $\mathcal{F}^t(\cdot)$ and fed to the stage GoMBF - $\{W_P^t, \Phi_P^t(\cdot)\}$. The extracted pixels should contain the discriminative information of facial motion and their locations should be invariant against similarity transform (scale, rotation and translation). To this end, we propose to generate the feature points by randomly sampling around the local regions of reference 2D landmarks (the mean of all training images' 2D landmarks) and index them by the barycentric coordinates with respect to the closest Delaunay triangles formed by those landmarks as in [2]. Before each stage regression, we first generate $M$ feature points and save the corresponding triangle indices and barycentric coordinates. Then, $\mathcal{F}^t(\cdot)$ calculates 2D landmark positions from the previous facial shape estimation - $\{\boldsymbol{\alpha}, \mathbf{Q}, P^{t-1}\}$ and calls the saved indexing information to extract pixels from the image.

## V. EXPERIMENTS

This section first validates the proposed GoMBF and GoMBF-Cascade regression in 3D facial tracking on in-the-wild videos. Then it deeply investigates the effect of various synthetic data on training GoMBF-Cascade for 3D facial tracking.

***Implementation.*** The GoMBF-Cascade regression involves a pack of parameters which we set as follows: offline training - $T = 10$, $F = 5$, $K_{\boldsymbol{\delta}} = K_{\boldsymbol{\theta}} = K_{\mathbf{t}} = K_{\mathbf{D}} = 80$, $M = 600$; runtime prediction - $L = 20$. This parameter configuration applies to all the following experiments without further specification. Since this work focuses on accurate facial motion regression, there is no post-processing and parametric face model adaption during online tracking as in previous studies [2][8]. The focal length $f$ is empirically set as 1,000 and the facial identity coefficients $\alpha$ are estimated from the first video frame and keep fixed for the rest frames. Whereas the setup is somewhat rough and poses much bigger challenges on facial motion regression, our GoMBF-Cascade is able to produce accurate and temporally-smooth tracking results. The tracking system is implemented using C++ with OpenMP parallelization, and tested on a laptop with a quad-core Intel Core i5 (2.30GHz) CPU and an integrated web camera producing 640 x 480 video frames. The system achieves a 30fps performance.

TABLE I
TRAINING AND TESTING DATASETS

| | Training Set |
|---|---|
| 300W-3D | 3,837 images, >500 subjects |
| FaceWarehouse | 1,600 images, 80 subjects |
| Multi-PIE | 1,024 images, 63 subjects |
| | Testing Set |
| 300VW | 004, 007, 009, 018, 019, 028, 037, 044, 048, 119, 143, 205, 208, 213, 223, 224, 405, 524, 531, 558. |
| Live Video Streams | |

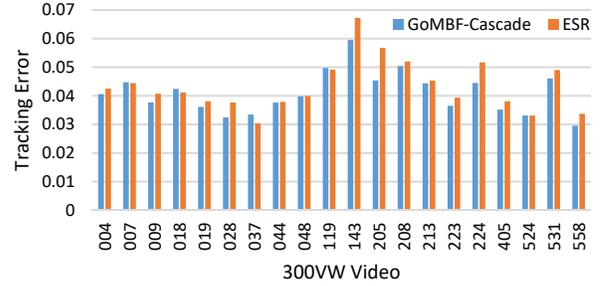

Fig. 4. Comparison between GoMBF-Cascade and ESR on 2D landmark tracking.

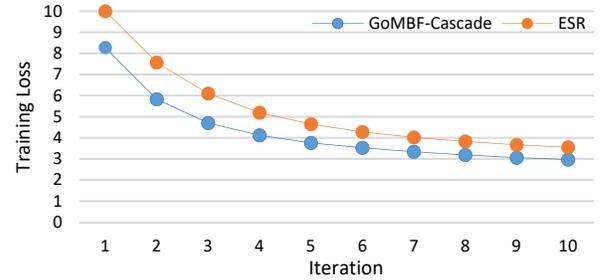

Fig. 5. The training convergence curves of GoMBF-Cascade and ESR.

*A. GoMBF-Cascade Validation*

*1) Datasets*

**Training Data.** Ideally, the proposed method requires an image dataset with accurate 2D landmark annotations and ground-truth 3D shape parameters that match our parametric face model for training. However, there is no such data available. As an alternative, we select images from three public face datasets and generate the corresponding 2D/3D labels by ourselves. Our training images are from 300W-3D [35], FaceWarehouse [26] and Multi-PIE [36]:

*300W-3D* contains 3,837 in-the-wild face images, each being offered with 68 hand-labelled landmarks (we discard the two points on the inner mouth corners in this work) and a reconstructed 3D facial mesh. For each image, we first estimate identity and expression coefficients - $\{\alpha, \delta\}$ by fitting our parametric face model to the provided 3D facial mesh based on landmark constraints. The fitting process resembles the one expressed in Eq. (10) with the only difference that $E_{lan}$ measures 3D landmark distances in world space this time. We then get $\mathbf{Q}$ by fixing the focal length $f$ to 1,000 and estimate the rotation and translation parameters - $\{\theta, t\}$ using the POSIT algorithm [33]. Finally, $\mathbf{D}$ can be easily calculated by comparing the 2D landmarks projected from the estimated 3D face to the hand-labelled landmarks.

*FaceWarehouse* consists of 3,000 near-frontal face images captured from 150 human subjects under controlled indoor environment. We choose 1,600 images of 80 subjects to use in our experiment and detect 66 landmarks for each image using a pre-trained SDM model [31]. Since the algorithmic landmark detection is not accurate enough, we go through all the images and manually adjust the misaligned landmarks. We later follow the process explained in Eq. (10) to estimate 3D shape parameters from 2D landmark labels. To correct implausible facial expression estimations, we further manually tune the expression coefficients. The identity and head pose parameters are updated afterwards to align with the new facial expression coefficients using a similar fitting method as described above.

*Multi-PIE* provides more than 4K indoor face images captured from 337 subjects. The images cover various facial expressions, head poses and illumination conditions. Each image has been manually annotated with 68 landmarks. We select 1,024 images of 63 subjects for training our facial motion regression. The corresponding 3D facial data is obtained with the same approach used for processing the FaceWarehouse data.

Overall, we collected 6,461 images for training. Table I shows the basic information of the training set. Despite the relatively smaller size of training set, the proposed GoMBF-Cascade regression delivers tracking results competitive to the state-of-the-art method [11] that used much more training data.

**Testing Data.** The tracking system has been evaluated on 20 challenging in-the-wild videos from 300VW [37]. Each video records the facial performance of a human subject in an unconstrained environment. The videos have been labelled with 68 2D landmarks frame by frame, providing a good benchmark to assess our tracking system that also outputs 2D landmarks. After scrutinizing the videos, we discard those that cannot be tracked since the first frame and then randomly select 20 videos from the rest of 300VW. The corresponding video information is listed in Table I. In addition, our tracking system has also been tested on live video streams.

*2) Comparison with State-of-the-art Methods*

We first validate GoMBF by comparing GoMBF-Cascade

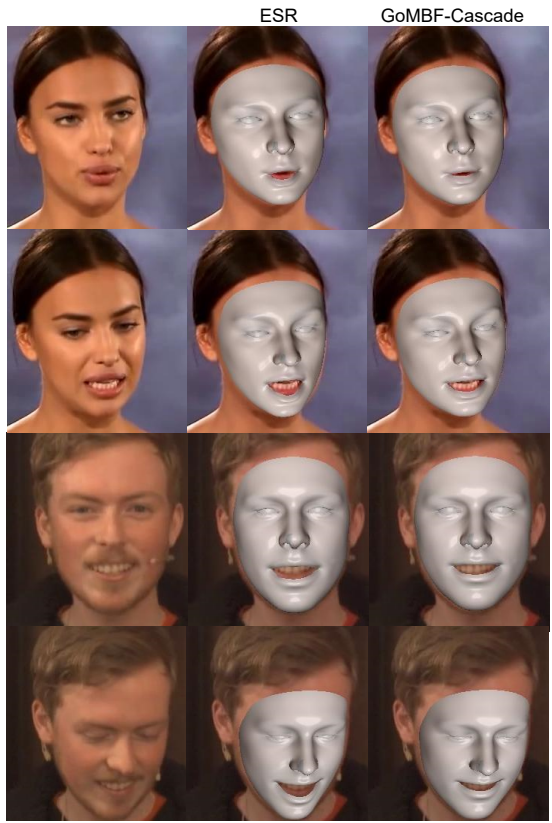

Fig. 6. GoMBF-Cascade tracks facial expressions more accurately than ESR.

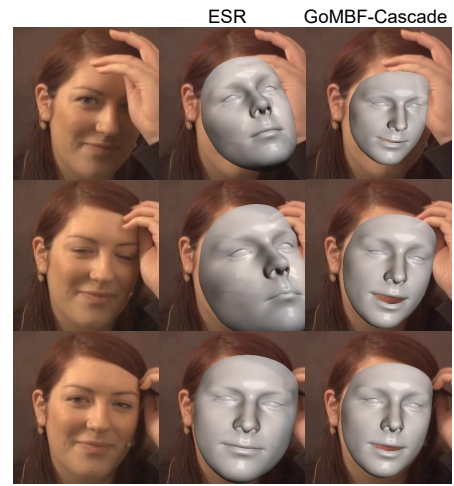

Fig. 7. GoMBF-Cascade shows higher resilience to occlusions than ESR.

with the Explicit Shape Regression (ESR) method [15] which has been widely applied in 2D/3D facial tracking [1][2][6]. GoMBF-Cascade differs from ESR by employing GoMBF instead of the conventional boosted ferns as the stage regressor. For fair comparison, we implement ESR with the same appearance vector extraction function $\mathcal{F}(\cdot)$ and other setups as that used in GoMBF-Cascade regression: training - $T = 10$, $F = 5$, $K = 320$, $M = 600$; runtime prediction - $L = 20$. ESR and GoMBF-Cascade are then trained on the same training set as introduced above. We test the two tracking models - GoMBF-Cascade and ESR on the selected 300VW videos. The tracking results are evaluated both quantitatively and visually.

For quantitative comparison, we apply the widely-accepted point-to-point root mean square error (normalized by the face's inter-ocular distance) between the tracked 2D landmarks and the ground-truth annotations [37]. For each video, we report the error averaged over all landmarks and video frames. As shown in Fig. 4, GoMBF-Cascade delivers lower 2D landmark tracking error for most of the 300VW videos than ESR. In addition to the improved runtime tracking performance, GoMBF-Cascade exhibits faster convergence than ESR during training (see Fig. 5, the error is measured as the RMSE between the 2D landmark annotations and predictions). We believe this benefits from the mechanism of GoMBF which exploits the complementary information between inter-modular boosted ferns and refines fern leaves towards the final regression target. What's more, GoMBF-Cascade enables parallel training when learning the modular boosted ferns. This significantly reduces the training time as compared to the traditional boosted ferns which has to be learned sequentially. In our experiment, it took about 3,532s to train a stage boosted ferns in ESR, while it only took 1,228s for GoMBF-Cascade, saving about 65.2% training time. We also visually compare GoMBF-Cascade and ESR by rendering out the tracked 3D faces. As shown in Fig. 6, GoMBF-Cascade is able to track facial expressions especially the mouth movements more precisely than ESR. GoMBF-Cascade is also found to be much more resilient to occlusions than ESR (see Fig. 7). In conclusion, both quantitative and visual results demonstrate that the proposed GoMBF outperforms the conventional boosted ferns in fitting power and learning speed.

To further verify the robustness of our GoMBF-Cascade regression, we compare its tracking results with those output from two state-of-the-art 3D facial tracking approaches - [10] and [11]. [10] is a typical optimization-based method which casts the tracking process into minimizing a highly non-linear objective function that enforces alignment on sparse feature points and pixel intensities between the reconstructed 3D face and the input video frame. It relies on GPU computing to achieve real-time performance. [11] instead resorts to convolutional neural networks (CNN) to regress facial shape and appearance parameters from facial images. It used 80K facial images to train its tracking network. As shown in Fig. 8, GoMBF-Cascade achieves competitive tracking performance against the two methods that either relied on a complicated photo-geometric fitting process [10] or was trained on a large-scale dataset [11]. It does better on tracking eye closure and mouth movements than those two methods. Furthermore, tracking results on live video streams also demonstrate the robustness of GoMBF-Cascade (see Fig. 9). Please note that our tracking results are purely based on the proposed GoMBF-Cascade regression without any post-processing on the regressed expression and head pose parameters. As demonstrated, GoMBF-Cascade provides a robust and elegant solution to 3D facial tracking with a reasonably small set of training data. (For more tracking results, please refer to the supplementary video)

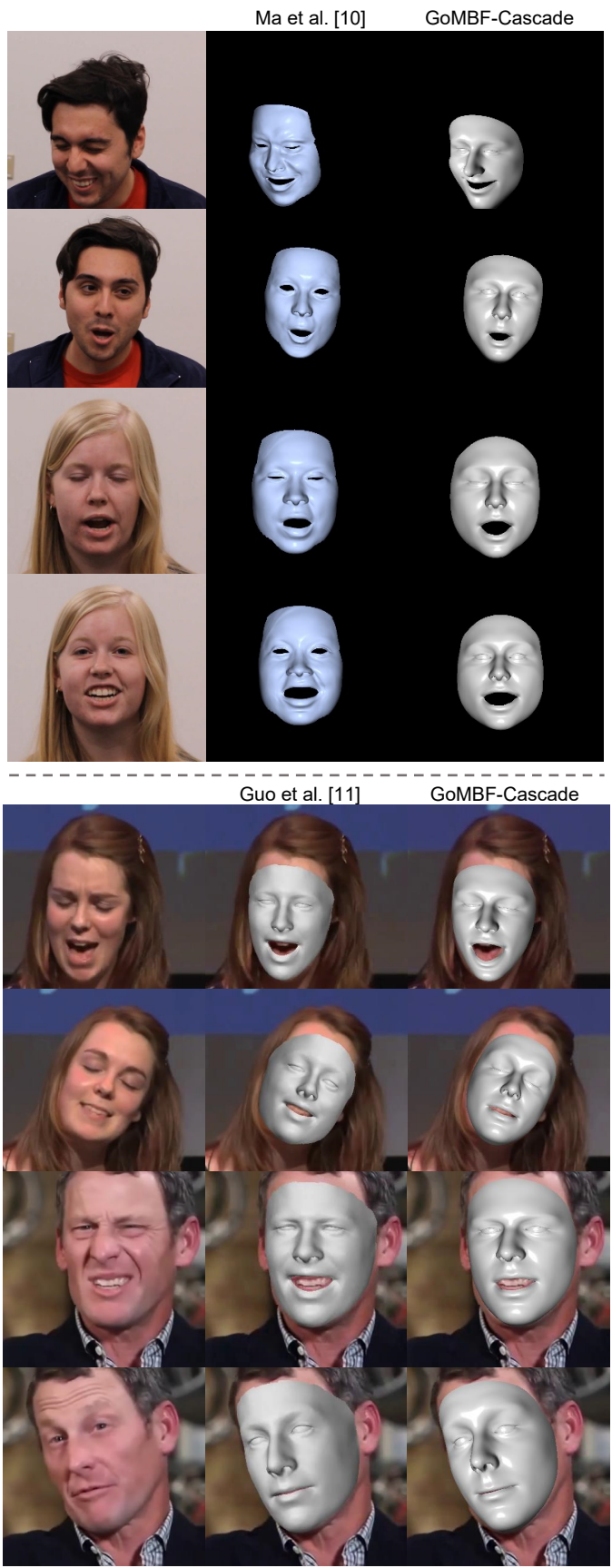

Fig. 8. Comparison between GoMBF-Cascade and [10] and [11]. Please note that the authors of [11] provided us the videos for testing and comparison.

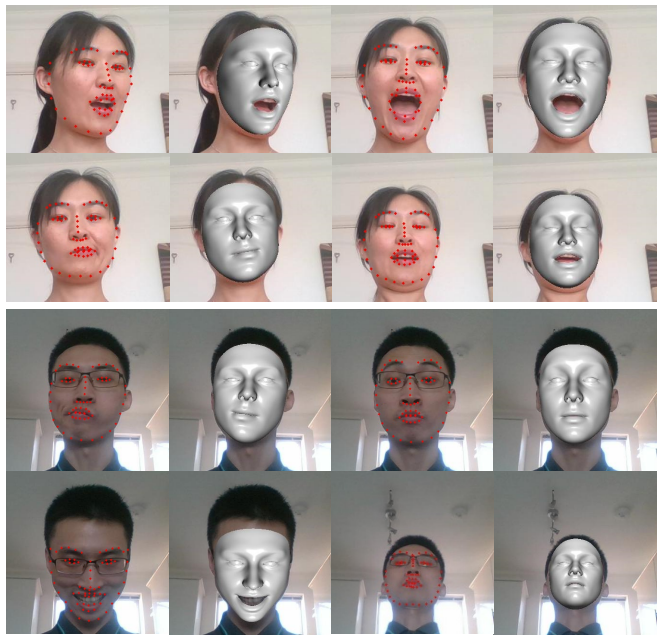

Fig. 9. Tracking results of GoMBF-Cascade on live video streams.

### B. Training with Synthetic Data

As described above, collecting facial images with accurate 3D geometry is tedious, which normally needs time-consuming human inspection and correction. As an alternative, synthesizing facial imagery for training is highly-efficient and provides fully accurate 3D labels. Whereas this novel data harvesting method has been successfully applied in 3D facial tracking and reconstruction using deep learning [11][19][20], it remains unclear if the synthetic data also favours non-deep learning methods such as GoMBF-Cascade. This part investigates the largely unexplored problem by using three types of synthetic facial imagery with various naturalness levels to train GoMBF-Cascade. The tracking models trained on real data, on synthetic data and on a mixture of data are then compared with each other.

#### 1) Synthesizing Training Imagery

In computer graphics, simulating real-world lighting and facial texture is crucial in rendering photo-realistic faces. Based on this insight, we apply three different lighting and texture models to synthesize facial imageries with various naturalness levels:

At the first stage, we incorporate BFM's texture components [23] and Phong illumination [24] into our parametric face model to render new faces. To cover a wide range of facial shapes, poses and lighting conditions, we construct multiple groups of rendering parameters. Specifically, we generate 40 3D heads by randomly sampling shape and texture coefficients from the corresponding normal distributions provided by BFM. For each head, 30 samples in various poses are generated, including 10 with neutral expression and specified head poses, 10 in frontal pose but with specified expressions, and 10 with random head poses and expressions (head pose and expression coefficients are chosen from the pre-built 300W-3D dataset).

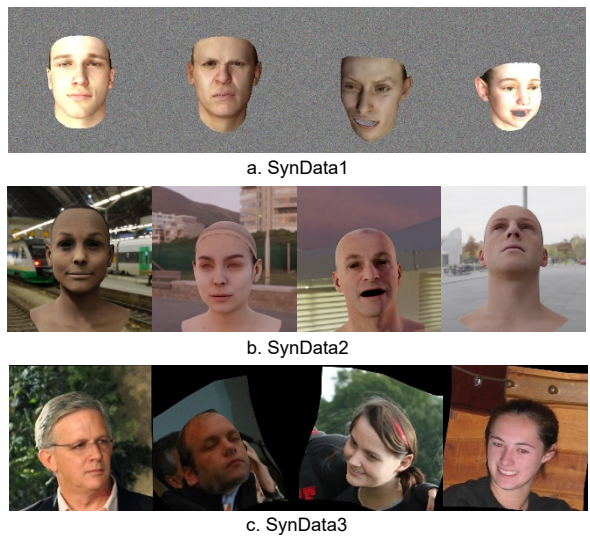

Fig. 10. Synthesized facial images.

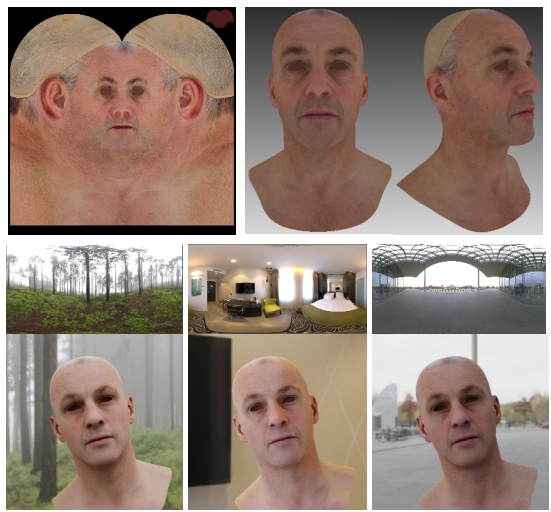

Fig. 11. Materials for generating SynData2. Top row: high-quality texture map and textured 3D facial mesh; middle row: HDR images; bottom row: the synthesized facial images.

TABLE II
SYNTHETIC DATASET

| | |
|---|---|
| SynData1 (4,800 images) | - 40 3D heads randomly generated from BFM.<br>- Each head is rendered with 30 different poses and 4 Phong illumination conditions. |
| SynData2 (4,800 images) | - 40 3D heads with very high-quality texture.<br>- Each head is rendered with 30 different poses and 4 natural lighting conditions simulated with HDR images. |
| SynData3 (9,300 images) | - Selected from CoarseData [13] which was built by applying lighting and texture estimated from in-the-wild facial images. |

To render each head sample, we randomly select four lighting conditions from a set consisting of 72 Phong illumination models which vary in components of direction, specular reflection, diffuse reflection, ambient reflection and shininess. In total, we synthesize 4,800 3D head samples. After defining a perfect pinhole camera model with a focal length of 1,000 and setting the image size to 450 x 450, all the 3D heads are rendered to images with a background filled with Gaussian noise (see Fig. 10a for examples).

As BFM [23] and Phong illumination [24] only simulate the facial texture and scene lighting in a coarse level, the synthetic faces from the first stage look rough and present clear artefacts. To improve the naturalness of the synthesized faces, we utilise a bundle of 20 high-quality head texture maps captured with a commercial photogrammetry rig[1]. The texture maps pair with two base meshes of an identical topology. Each texture occupies exactly the same UV and can be swapped out conveniently for a different texture, hence resulting in 40 different 3D heads. Since the base mesh is in repose, we generate its delta blendshapes using deformation transfer [27] to enable facial expression modelling. For each base mesh, we also manually annotate 66 landmarks which share the same semantic meaning as those used in our parametric face model. With the matched 3D landmarks, we can easily estimate from the synthesized 3D head the required shape parameters using the approach mentioned in 300W-3D data processing. To realistically illuminate the head, we apply an image-based lighting technique in which high dynamic range (HDR) panoramic images are used to provide the environment lighting. The technique captures omni-directional light information of a real-world scene and stores it into pixels of a HDR image which can be projected to a sphere simulating the surrounding space of the target object. We collect 12 HDR images (see Fig. 11) which were captured from common indoor and outdoor scenes such as train station, hotel room and misty pines. Following the procedure as described in the first stage, we generate 30 samples in various poses for each 3D head. Each head sample is then rendered to 450 x 450 images with four lighting conditions and the image's background is set as the scene exhibited in the corresponding HDR image for more natural synthesis. We use the inbuilt Cycles path-tracing engine of Blender[2] for rendering. As shown in Fig. 10b, highly photo-realistic facial images with fine texture features such as pores and wrinkles can be synthesized.

Comparing to in-the-wild data, facial images synthesized in the first two stages still have pronounced artefacts, e.g. the lack of inner-mouth structure, limited variations in facial shape, lighting and background. In this stage, we turn to another kind of synthetic data [11] which is derived from using facial shape, texture and lighting estimated from real-world images for rendering. By further warping the background region of the source image to fit the new face, the synthetic image can look very similar to real-world counterpart. Guo et al. [11] have released such a dataset which was named CoarseData (see Fig. 10c). The dataset was generated from 3,131 300W-3D images, with each image being augmented 30 times to cover more facial expressions and head poses. In our experiment, we randomly select 9,300 images (about 3 samples for each original 300W-3D image) from CoarseData and apply the same method used in processing the 300W-3D data to get the ground-truth shape parameters that fit our parametric face model.

---

[1] https://www.3dscanstore.com/.

[2] https://www.blender.org/.

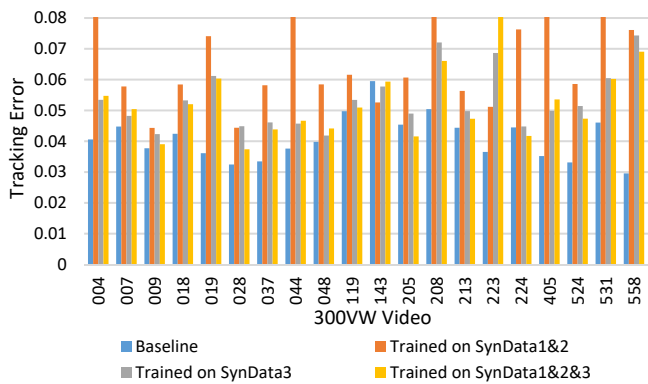

Fig. 12. Comparison between the tracking models trained purely on synthetic data and the baseline model (error ≫ 0.08 has been cut off).

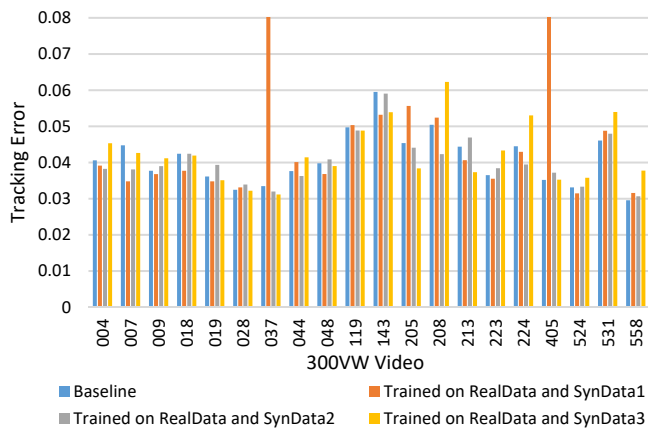

Fig. 13. Comparison between the tracking models trained on the mixture of data and the baseline model (error ≫ 0.08 has been cut off).

For convenience, we call the training set built in *Part A* as RealData, the three synthetic datasets as SynData1, SynData2 and SynData3 respectively. The corresponding information is listed in Table II.

*2) Tracking Model Comparison*

Taking the GoMBF-Cascade regression trained on RealData as the baseline model, we evaluate the models trained purely on synthetic data or on a mixture of real and synthetic data in tracking 300VW videos.

***Training on synthetic data.*** To introduce sufficient facial shape and appearance variations during regression learning, we train three GoMBF-Cascade models which are on the mixture of SynData1 and SynData2 (SynData1&2), on SynData3, and on all the synthesized facial images (SynData1&2&3) respectively. We follow the same setups as training with RealData. The three models are then tested on 300VW videos and compared to the baseline model by calculating the aforementioned 2D landmark tracking error. As shown in Fig. 12, the models trained purely with synthetic facial images output much bigger tracking errors than the baseline model, especially the model trained with SynData1&2 which completely lost the face in some videos such as video-044 (with an error of 110.47). Even for the models trained with SynData3 which comprises synthesized facial images looking very close to the real in-the-wild data, the tracking accuracy is still sharply lower than the baseline model's.

It is worth pointing out that McDonagh et al. [6] successfully learned an ESR-based facial motion regression for personalized 3D facial tracking from synthesized training imageries. However, their synthetic images were rendered from a high-quality facial rig (built from an offline facial capture system) that fits tightly with the user's facial geometry and appearance, and an illumination model driven by light probe data acquired at the target environment. This can hardly be achieved in unconstrained facial tracking scenario where the target environment and user are unknown in the training phase.

***Training on mixed data.*** To further investigate the impact of synthesized facial images, we sequentially mix the synthetic data with the real data for training GoMBF-Cascade. As a result, we generate three tracking models. Fig. 13 presents the 2D landmark tracking errors of these three models on 300VW videos. As shown in the figure, for most videos, at least one of the three models trained on the mixed data exhibits improved tracking performance than the baseline model. However, none of them is as reliable as the baseline model that can be generalized well to all the testing videos. The model trained with SynData1 even outputs an extremely large tracking error – 19.56 on video-037.

From these two experiments, we find that: 1) the GoMBF-Cascade tracking model trained purely on synthesized facial images cannot generalize well to unconstrained real-world data; 2) involving synthetic facial images into training benefits tracking in some certain scenarios, but degrades the tracking model's generalization ability. Interestingly, these two findings are contrary to those observed in deep learning-based 2D/3D facial tracking and reconstruction [11][19][20][25]. As reported in [11][19][20], facial tracking/reconstruction CNN models that work well on real-world data can be learned only with similar synthetic facial images as SynData1 and SynData3. In [25], the authors find that priming deep networks by pre-training them with synthetic facial images is helpful for reducing the negative effects of the training data bias. Presumably, the discrepancy is mainly caused by the different feature learning capabilities between deep and non-deep learning methods.

## VI. CONCLUSION

In this paper, we first develop a novel regression method called GoMBF-Cascade for real-time 3D facial tracking from a monocular RGB video. GoMBF-Cascade is mainly featured with a sequence of globally-optimized modular boosted ferns – GoMBF, which is built with compositional learning and can efficiently handle the modality variety in facial motion parameters during regression. Compared with the conventional boosted ferns [1][2][15], GoMBF exhibits stronger fitting power and a higher learning speed. In theory, GoMBF can be seamlessly adapted to any other multi-output regression tasks. The resulting GoMBF-Cascade regression has been validated in 3D facial tracking on in-the-wild videos and live video streams. It delivers competitive tracking performance comparing against the state-of-the-art methods [10][11] which require a large-scale training set or have a much higher computational complexity, hence providing a robust and highly elegant solution to real-time 3D facial tracking.

We also systematically investigate the effect of synthesized facial imageries on training GoMBF-Cascade for 3D facial tracking. We apply three different kinds of synthetic facial images with various naturalness level for training and compare the tracking models trained on real data, on synthetic data and on a mixture of data. Our experimental results show that, 1) training purely with synthesized facial images can hardly deliver a robust 3D facial tracking model that generalizes well to unconstrained real-world data; 2) involving synthetic images into training can benefit tracking in some certain scenarios, but harms the tracking model's generalization ability. This provides a different understanding of learning from synthetic facial images as those formed in deep learning-based 2D/3D facial tracking and reconstruction [11][19][20][25]. It is supposed to be caused by the different feature learning capabilities between deep and non-deep learning approaches. We believe the findings can benefit a series of non-deep learning facial image analysis tasks where the labelled real data is difficult to access.

We notice that, by conditioning on facial pose, expression and illumination, the generative adversarial network (GAN) [38] is able to synthesize extremely realistic facial images. This provides a highly flexible and efficient way for synthesizing facial imagery with ground-truth labels. In the future, it would be a very promising direction to apply such networks to generate the training data for 3D facial tracking.

REFERENCES

[1] C. Cao, Y. Weng, S. Lin, and K. Zhou, "3D shape regression for real-time facial animation," *ACM Trans. Graph.*, vol. 32, no. 4, pp. 1-10, 2013.
[2] C. Cao, Q. Hou, and K. Zhou, "Displaced dynamic expression regression for real-time facial tracking and animation," *ACM Trans. Graph.*, vol. 33, no. 4, pp. 1-10, 2014.
[3] C. Cao, D. Bradley, K. Zhou, and T. Beeler, "Real-time high-fidelity facial performance capture," *ACM Trans. Graph.*, vol. 34, no. 4, pp. 1-9, 2015.
[4] J. Thies, M. Zollhöfer, M. Stamminger, C. Theobalt, and M. Nießner, "Face2face: Real-time face capture and reenactment of rgb videos," in *Proc. IEEE CVPR*, Las Vegas, NV, USA, 2016, pp. 2387-2395.
[5] H. Chen, J. Li, F. Zhang, Y. Li, and H. Wang, "3D model-based continuous emotion recognition," in *Proc. IEEE CVPR*, Boston, MA, USA, 2015, pp. 1836-1845.
[6] S. McDonagh, M. Klaudiny, D. Bradley, T. Beeler, I. Matthews, and K. Mitchell, "Synthetic prior design for real-time face tracking," in *Proc. IEEE 3DV*, Stanford, CA, USA, 2016, pp. 639-648.
[7] C. Wang, F. Shi, S. Xia, and J. Chai, "Realtime 3D Eye Gaze Animation Using a Single RGB Camera," *ACM Trans. Graph.*, vol. 35, no. 4, pp. 1-14, 2016.
[8] S. Saito, T. Li, and H. Li, "Real-time facial segmentation and performance capture from rgb input," in *Proc. ECCV*, Amsterdam, Netherlands, 2016, pp. 244-261.
[9] S. Laine, T. Karras, T. Aila, A. Herva, S. Saito, R. Yu, H. Li, and J. Lehtinen, "Production-level facial performance capture using deep convolutional neural networks," in *Proc. ACM SCA*, Los Angeles, CA, USA, 2017, pp. 1-10.
[10] L. Ma and Z. Deng, "Real-time hierarchical facial performance capture," in *Proc. ACM I3D*, Montreal, QC, Canada, 2019, pp. 1-10.
[11] Y. Guo, J. Zhang, J. Cai, B. Jiang, and J. Zheng, "CNN-based real-time dense face reconstruction with inverse-rendered photo-realistic face images," *IEEE Trans. Pattern Anal. Mach. Intell.*, vol. 41, no. 6, pp. 1294-1307, 2019.
[12] J. S. Yoon, T. Shiratori, S. I. Yu, and H. S. Park, "Self-supervised adaptation of high-fidelity face models for monocular performance tracking," in *Proc. IEEE CVPR*, Long Beach, CA, USA, 2019, pp. 4601-4609.
[13] Y. Guo, J. Zhang, L. Cai, J. Cai, and J. Zheng, "Self-supervised CNN for Unconstrained 3D Facial Performance Capture from an RGB-D Camera," arXiv preprint arXiv:1808.05323, 2018.
[14] Y. Weng, C. Cao, Q. Hou, and K. Zhou, "Real-time facial animation on mobile devices," *Graphical Models*, vol. 76, no. 3, pp. 172-179, 2014.
[15] X. Cao, Y. Wei, F. Wen, and J. Sun, "Face alignment by explicit shape regression," *Int. J. Comput. Vis.*, vol. 107, no. 2, pp. 177-190, 2014.
[16] A. E. Hoerl and R. W. Kennard, "Ridge regression: biased estimation for nonorthogonal problems," *Technometrics*, vol. 12, no. 1, pp. 55-67, 1970.
[17] P. Dollár, P. Welinder, and P. Perona, "Cascaded pose regression," in *Proc. IEEE CVPR*, San Francisco, CA, USA, 2010, pp. 1078-1085.
[18] Dimensional Imaging, DI4D PRO System, 2016. [Online]. Available: http://www.di4d.com/systems/di4d-pro-system/.
[19] E. Richardson, M. Sela, and R. Kimmel, "3D face reconstruction by learning from synthetic data," in *Proc. IEEE 3DA*, Stanford, CA, USA, 2016, pp. 460-469.
[20] E. Richardson, M. Sela, R. Or-El, and R. Kimmel, "Learning detailed face reconstruction from a single image," in *Proc. IEEE CVPR*, Honolulu, HI, USA, 2017, pp. 1259-1268.
[21] M. Zollhöfer, J. Thies, P. Garrido, D. Bradley, T. Beeler, P. Pérez, M. Stamminger, M. Nießner, and C. Theobalt, "State of the art on monocular 3D face reconstruction, tracking, and applications," *Computer Graphics Forum*, vol. 37, no. 2, pp. 523-550. 2018.
[22] L. A. Jeni, J. F. Cohn, and T. Kanade, "Dense 3D face alignment from 2D videos in real-time," in *Proc. IEEE FG*, Ljubljana, Slovenia, 2015, pp. 1-8.
[23] O. Paysan, R. Knothe, B. Amberg, S. Romdhani, and T. Vetter, "A 3D face model for pose and illumination invariant face recognition," in *Proc. IEEE Int. Conf. Adv. Vid. Sig. Based Sur.*, Genova, Italy, 2009, pp. 296-301.
[24] B. T. Phong, "Illumination for computer generated pictures," *Communications of the ACM*, vol. 18, no. 6, pp. 311–317, 1975.
[25] A. Kortylewski, B. Egger, A. Morel-Forster, A. Schneider, T. Gerig, C. Blumer, C. Reyneke, and T. Vetter, "Can Synthetic Faces Undo the Damage of Dataset Bias to Face Recognition and Facial Landmark Detection?," arXiv preprint arXiv:1811.08565, 2018.
[26] C. Cao, Y. Weng, S. Zhou, Y. Tong, and K. Zhou, "Facewarehouse: A 3d facial expression database for visual computing," *IEEE Trans. Vis. Comp. Graph.*, vol. 20, no. 3, pp. 413-425, 2013.
[27] R. W. Sumner and J. Popović, "Deformation transfer for triangle meshes," *ACM Trans. Graph.*, vol. 23, no. 3, pp. 399-405, 2004.
[28] S. Ren, X. Cao, Y. Wei, and J. Sun, "Face alignment at 3000 fps via regressing local binary features," in *Proc. IEEE CVPR*, Columbus, OH, USA, 2014, pp. 1685-1692.
[29] S. Ren, X. Cao, Y. Wei, and J. Sun, "Global refinement of random forest," in *Proc. IEEE CVPR*, Boston, MA, USA, 2015, pp. 723-730.
[30] P. Viola and M. J. Jones, "Robust real-time face detection," *Int. J. Comput. Vis.*, vol. 57, no. 2, pp. 137-154, 2004.
[31] X. Xiong, and F. D. l. Torre, "Supervised descent method and its applications to face alignment," in *Proc. IEEE CVPR*, Portland, OR, USA, 2013, pp. 532-539.
[32] R. H. Byrd, P. Lu, J. Nocedal, and C. Zhu, "A limited memory algorithm for bound constrained optimization," *SIAM J. Sci. Comput.*, vol. 16, no. 5, pp. 1190-1208, 1995.
[33] D. F. Dementhon and L. S. Davis, "Model-based object pose in 25 lines of code," *Int. J. Comput. Vis.*, vol. 15, no. 1-2, pp. 123-141, 1995.
[34] X. Zhu, Z. Lei, J. Yan, D. Yi, and S. Z. Li, "High-fidelity pose and expression normalization for face recognition in the wild," in *Proc. IEEE CVPR*, Boston, MA, USA, 2015, pp. 787-796.
[35] X. Zhu, Z. Lei, X. Liu, H. Shi, and S. Z. Li, "Face alignment across large poses: A 3d solution," in *Proc. IEEE CVPR*, Las Vegas, NV, USA, 2016, pp. 146-155.
[36] R. Gross, I. Matthews, J. Cohn, T. Kanade, and S. Baker, "Multi-pie," *Image and Vision Computing*, vol. 28, no. 5, pp. 807-813, 2010.
[37] J. Shen, S. Zafeiriou, G. S. Chrysos, J. Kossaifi, G. Tzimiropoulos, and M. Pantic, "The first facial landmark tracking in-the-wild challenge: Benchmark and results," in *Proc. IEEE ICCVW*, 2015.
[38] A. Tewari, M. Elgharib, G. Bharaj, F. Bernard, H.-P. Seidel, P. Pérez, M. Zollhofer, and C. Theobalt, "StyleRig: rigging styleGAN for 3d control over portrait images," in *Proc. IEEE CVPR*, 2020, pp. 6142-6151.